# The Utility of Clustering in Prediction Tasks

Shubhendu Trivedi, Zachary A. Pardos and Neil T. Heffernan

*Abstract*—**We explore the utility of clustering in reducing error in various prediction tasks. Previous work has hinted at the improvement in prediction accuracy attributed to clustering algorithms if used to pre-process the data. In this work we more deeply investigate the direct utility of using clustering to improve prediction accuracy and provide explanations for why this may be so. We look at a number of datasets, run k-means at different scales and for each scale we train predictors. This produces k sets of predictions. These predictions are then combined by a naïve ensemble. We observed that this use of a predictor in conjunction with clustering improved the prediction accuracy in most datasets. We believe this indicates the predictive utility of exploiting structure in the data and the data compression handed over by clustering. We also found that using this method improves upon the prediction of even a Random Forests predictor which suggests this method is providing a novel, and useful source of variance in the prediction process.**

## I. Introduction

One of the motivations to this work is one of the author's (Zachary A. Pardos) successful participation in the 2010 KDD Cup, which involved a prediction task on an educational dataset. Methods such as Bagged Decision Trees were used to get the second position in the student category. The dataset had instances for a number of students. Since students can be crudely binned into categories in terms of learning rate, forgetting rate etc., a natural question to ask is if clustering the students and trying to find such groups would aid in classification accuracy. This question was crudely tested in the 2010 UCSD Data Mining competition (an e-commerce task) in which the fourth position was secured using this clustering method alone. Motivated by the success of this technique, an internal graduate Machine Learning course competition was organized at Worcester Polytechnic Institute (WPI) that explored this notion further. This idea of using clustering coupled with simple predictors beat more complex methods such as Support Vector Machines and Random Forests on the KDD cup development set. This also led to papers [1] [2] that explored this idea in an educational dataset. This paper essentially develops this notion further. The rest of the article is organized as follows: Section II reviews some work on clustering, such as a theoretical justification of using clustering for a classification task. Section III discusses the idea of using clustering in conjunction with a predictor in more detail, with section III A providing some more work and intuition on how we can use clustering to improve accuracy on a prediction task. Section IV talks of the empirical study carried out and section V gives an overview of the results obtained, section VI has a discussion of observations and open questions.

This work was supported in part by the National Science Foundation via grant "Graduates in K-12 Education" (GK-12) Fellowship, award number DGE0742503 and the Department of Education IES Math Centre for Mathematics and Cognition grant. Report Date: 05 September 2011

## II. Clustering

It is reasonable to say that at least some part of our understanding of the world is due to a semi-supervised process that involves some sort of clustering in a big way. An example would be our ability to tell, given a mixture of objects which are similar and belong to the same category. It has been suggested that a mathematically precise notion of clustering is important in the sense that it can help us solve problems at least approximately as solved by the brain [3]. Clustering is probably the most used exploratory data analysis technique across disciplines and is frequently employed to get an intuition about the structure of the data, for finding meaningful groups, also for feature extraction and summarizing. Given a space $X$, clustering can be thought of as a partitioning of this space into $K$ parts i.e. $f: X \rightarrow \{1, \dots, K\}$ This partitioning is done by optimizing some internal clustering criteria such as the intra-cluster distances etc. The value of $K$ is found usually by employing a second criterion that measures the robustness of the partitioning

While clustering is useful for data analysis and as a preprocessing step for a number of learning tasks, we are interested in the specific pre-processing task of using clustering to gain more information about the data to improve prediction accuracy. This leads to the questions: Can clustering of unlabeled data give any new information that can aid a classification task? It has been hinted in the literature that clustering of unlabeled data should help in a classification task as clustering can also be thought of as separating classes. It is not clear if clustering could help in a regression task, though there is some evidence [1][2]. Another question that could be asked is: Can a number of predictions obtained by varying clustering parameters give us access to new information that can be combined together to improve prediction accuracy even more? Can the idea of clustering as a predictor be formalized? Previous work comprehensively answers at least the third question. This is an important question to ask since the answer justifies using clustering in a prediction task. The next sub-section briefly discusses this work before proposing a simple scheme to utilize clustering in prediction.

### A. Related Work

One of the most basic results in Learning Theory is the Occam's Razor [4] i.e. if a set of m training examples can be described by a hypothesis using only $k \ll m$ bits, then we can be quite sure that the hypothesis generalizes well to unseen data. Another way of stating this is that compression implies learning for the description language of the hypothesis. If compression means learning then making predictions would mean decompression. The notion of *compression implies learning* for different description languages has lead to a number of important sample complexity bounds [5][6] and has been generalized to any description language by Blum & Langford [7]. This generalization, called the PAC-MDL bound gives a handle on understanding the generalization error and the tradeoff between good representations of the data and over-fitting it. Clustering too can be seen as a trade-off between the quality of the representing groups in the data and the complexity of the same.

The said PAC-MDL bound is defined for a transductive setting and essentially states that it is quite unlikely that a transductive classifier that does well on the training set will do badly on the test set. This can be formalized as follows: Consider we have a training set $S_{train}$ having $m$ labeled examples and a test set $S_{test}$ having $n$ unlabeled examples which are drawn independently from a distribution $D$. If $X$ is the instance and $X$ the target, then $S_{train} = \{X^m, Y^m\}$ and $S_{test} = \{X^n, Y^n\}$ with $Y \in \{1, \dots, l\}$. Given any compression procedure as discussed in the previous paragraph which could be represented as $A: (X \times Y)^m \times X^n \to \{0,1\}^*$ there would be a decompression procedure $B: X^{m+n} \times \{0,1\}^* \to Y^{m+n}$. For this compression-decompression pair the transmitted string $\sigma$ would be the transductive classifier $\sigma: X^{m+n} \to Y^{m+n}$ that assigns labels to the examples. For a description language the bound on the test error ($\hat{\sigma}_{test}$) as a function of the error on the training set is given by the PAC-MDL bound [7] [8]:

*For any given distribution $D$ and for the set of all description languages $L = \{\sigma\}$ with probability $1 - \delta$ over the train and test sets:*

$$S_{train}, S_{test} \sim D^{m+n}: \forall \sigma$$
$$\hat{\sigma}_{test} \leq bmax(m, n, \hat{\sigma}_{train}, 2^{-|\sigma|}\delta)$$

While the PAC-MDL bound is used for a transductive setting Banerjee & Langford [7] show that clustering can be converted to a transductive classification problem. They also demonstrate that for a description language $L = \{\sigma\}$ to be a *valid* description language, it must be an instantaneous code and hence satisfy Kraft's inequality. For the case of clustering, with $c$ clusters and $l$ labels, the family of descriptions $L = \{\sigma\}$ has size $l^c$. This set can be encoded by $|\sigma| = clog(l)$ bits. Since L satisfies Kraft's inequality, it is a valid description language. This essentially gives an information theoretic justification of using clustering as a transductive classifier and also gives a set of PAC-MDL bounds on the same.

The above review in simple terms states the following: Since clustering is a scheme for information compression. It will thus (when stated as a transductive problem for simplicity) most likely improve the prediction error. The PAC-MDL bounds that formalize this notion can be used without any loss of generality as an intuitive explanation of why clustering could be used in conjunction with a predictor as a pre-processing step. The next section returns to the notion of using clustering for prediction.

## III. Using Clustering for bootstrapping

Clustering is used to mine structure in the data. According to a pre-defined metric data-points in one group are by definition highly similar to each other than to data-points from other groups/clusters. One useful way of looking at this is thinking of clustering as [9]: Consider a dataset that is obtained by sampling a collection of distributions $\{D_1, D_2, \dots, D_k\}$ with associated weights $\{w_1, w_2, \dots, w_k\}$ such that $\sum_i w_i = 1$ i.e. from each distribution $D_i$, a point is picked with probability $w_i$. Now given the dataset, the idea behind clustering is to identify these distinct distributions that might have generated it and assign points in the dataset into different groups accordingly. This new representation is more concise.

Following from the above and from the discussion in section II: Given a dataset, clustering it gives a compressed representation (albeit lossy). This can be thought of as giving the data to an operator as input (k-means for example) that gives an output of the same data but taking much fewer bits to represent it. This transformation tells us something interesting about the data and its structure which could be exploited to improve the predictive power. One potential way of doing so is by training a separate predictor on each cluster rather than train a single predictor on the entire dataset.

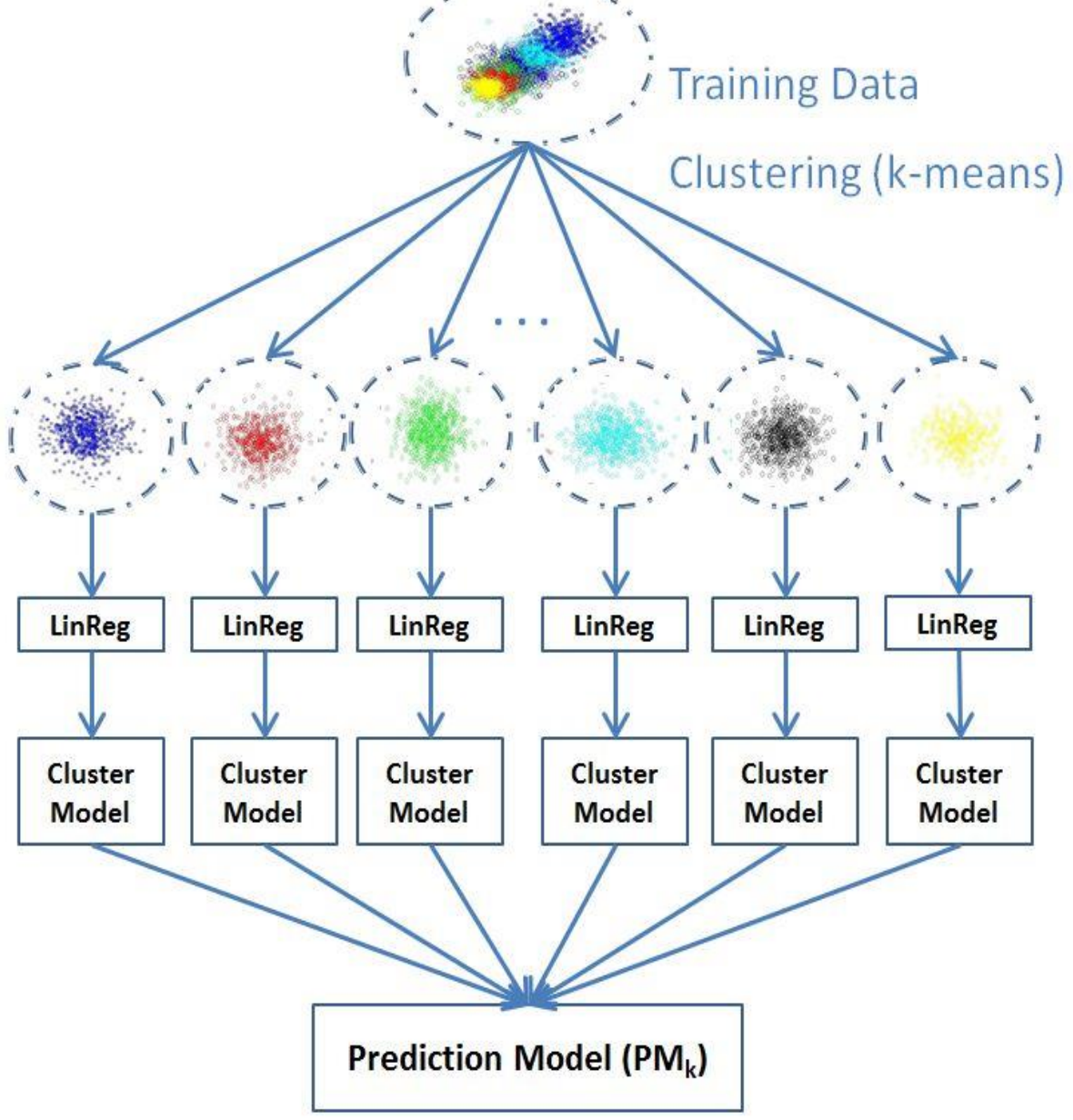

Fig. 1. A "Prediction Model". A "prediction model" is composed of k cluster models ($PM_k$). It should be noted that any other method for regression could be used in place of Linear Regression

Consider a sample regression task (Fig. 1): Suppose we first cluster the dataset into k clusters using an algorithm such as k-means. A separate linear regression model is then trained on each of these clusters (any other model can be used in place of linear regression). Let us call each such model a "Cluster Model". All of the k Cluster Models together can be thought of as forming a more complex model that we call a "*Prediction Model*". We represent a prediction model as $PM_k$, with the subscript indicating the number of cluster models in the given prediction model (which in turn will obviously equal the number of clusters). To summarize, to train a "prediction model", the following steps are followed:

1. Cluster the training data into k partitions
2. For each partition train a separate classifier/predictor using the points inside that cluster as its training set.
3. Each such predictor represents a model of the cluster, and hence is called the cluster model.

Once a prediction model is obtained, making a prediction of a point from the test set would involve the following (Fig. 2.)

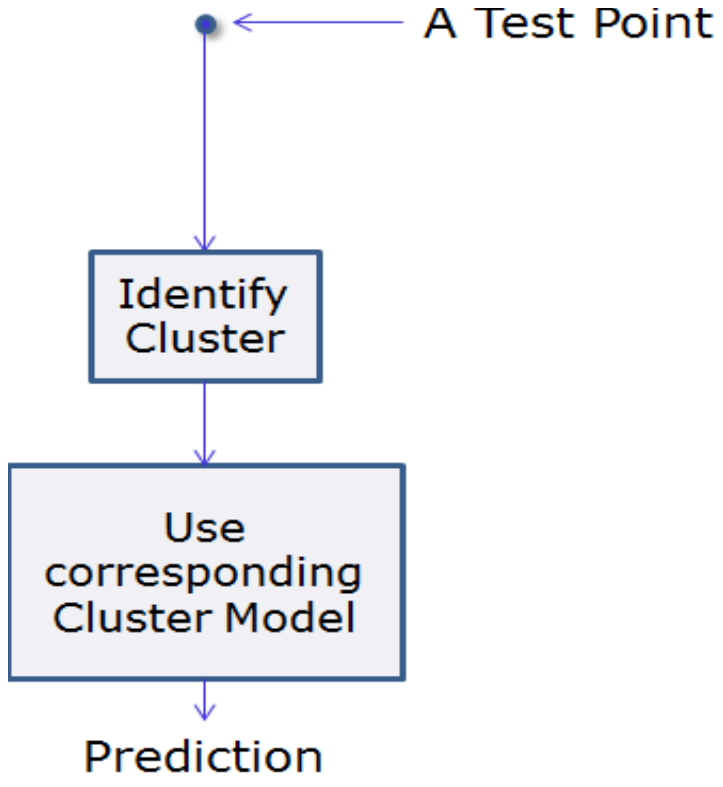

Fig. 2. Mapping a test point to a cluster to make a prediction on it

Making predictions for a point from the test set would thus involve two steps:

1. Identify the cluster to which the test point belongs.
2. Use the Cluster Model of the identified cluster to make the prediction for that data point.

It must be noted that $PM_1$ would simply be our predictor fit on the entire data set (for the above example it would be fitting a linear regression model on the dataset, we can think of the entire dataset as one cluster).

## A. *k as a tunable parameter*

The previous section describes a way by which clustering could be used to construct what we call a "prediction model". Building on the generic method, using the number of clusters 'k' in k-means (or any other clustering that requires number of clusters to be input) as a free parameter, multiple prediction models can be obtained (Fig. 3.) i.e. k can be varied from 1 to a value K and a Prediction Model for each instance can be obtained. For example if K = 3, there would be three prediction models: $PM_1$ (predictor trained on the entire dataset), $PM_2$ (predictors trained on two clusters), and $PM_3$ (predictors trained on three clusters). These K prediction models are then employed to make a set of K distinct predictions on the test set using the two step procedure for mapping and making predictions of test points sketched in the previous section. Before looking at how these K predictions can be of value, it must be noted that

1. Cluster models in different prediction models are different.
2. There might indeed exist a prediction model $PM_i$ for some arbitrary number of clusters $i$ that would have higher prediction accuracy than $PM_1$. The reverse might also be true.

The second factor i.e whether some arbitrary $PM_i$ would do better than $PM_1$ would depend on two main factors: Clusterabilty of the dataset [3] and the choice of predictor.

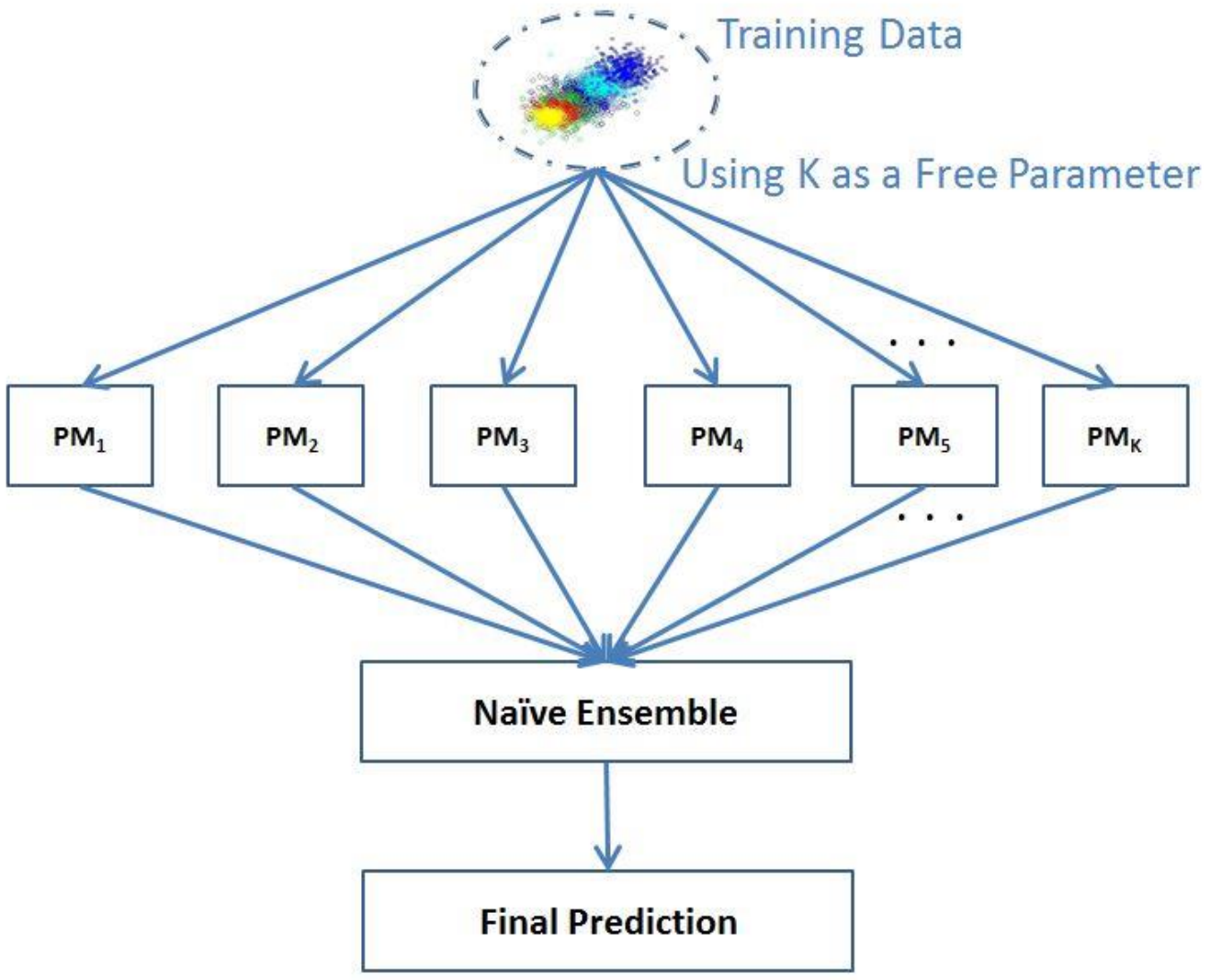

Fig. 3. Generation of multiple prediction models by using 'k' as a free parameter. Each of these prediction models will make a prediction on the test set. These predictions can then be combined together by a naïve ensemble to get a final prediction.

Even if an arbitrary $PM_i$ does not return higher accuracy than $PM_1$, a couple of questions of considerable interest would be: How good are the predictions made by each individual prediction model? How diverse are the predictions made by the various prediction models? If there is indeed some diversity in the error patterns in predictions made by the various prediction models, the next step would be to combine the predictions together to perhaps get a stronger prediction.

## B. *Combining Predictions*

Before looking at combining predictions, it is useful to understand how the predictions made by the various prediction models might be diverse and why combining diverse predictions might be helpful.

### *Information Theoretic View of Clustering*

As discussed in sections II and III, clustering seems useful for prediction as it is basically a scheme for data compression. By compression we learn something interesting about the structure and the regularities in the data that can be used to perhaps improve the prediction accuracy. A simple method to do so was outlined in section III-A. Interestingly however, how much compression we can achieve will depend on what 'k' (number of clusters) is chosen. A question that arises is: Is there at least some difference in the information content in these different cases? Let's consider this question in some detail: Consider k-means clustering; Now since the cluster centroids are found by optimizing a distortion function, the choice of this distortion function decides what information should be kept and what should not be. The distortion function for k-means is given by:

$$J(c,\mu) = \sum_{i=1}^{m} \|x^{(i)} - \mu_{c^{(i)}}\|^2$$

$\mu_c$ is the cluster centroid to which a point $x$ has been assigned.

The data are described more concisely (and hence the compression) with all the points in a cluster approximated by their corresponding cluster centroids found using the distortion function. The rest of the irrelevant data is thrown away

Previous work by Still & Bialek [3] has formalized and extended this notion of relevance. This formalization gives a tradeoff between the complexity of the model and the amount of relevant information. The tradeoff range gives an optimal number of clusters for a dataset of a finite size beyond which we begin to over-fit it. Other than this, the rate distortion theory applied to the problem of clustering also shows that the amount of relevant information coded at a certain clustering scale is different. Thus, in a sense there is no single best clustering of the data but a family of solutions that evolves with the tradeoff parameter [3]. This tradeoff in turn formalizes the notion of "clusterabilty" of the dataset and gives the valuable insight that at different values of this tradeoff we might get access to different information. While some of this information might be redundant, and some of it might be sampling noise, some information may also be unique to a grouping. We believe that it is this source of novel information that lends at least some power to the use of clustering in a prediction task. In our case, it would lend some diversity to the predictions obtained by the various prediction models since each is trained at a different scale of clustering.

*Ensemble Learning*

When we have a set of diverse (and accurate) predictors, combining them together to obtain a single prediction leads to ensemble methods (ensemble methods can also be considered methods as ways of generating diverse and accurate individual predictors in the first place).

Ensemble methods have seen a rapid growth in the past decade in the machine learning community [10][11][12]. An ensemble is a group of predictors each of which gives an estimate of a target variable. Ensemble learning is a way to combine these predictions with the goal that the generalization error of the combination is lesser than each of the individual predictors. The success of ensembling lies in the ability to exploit (or inject and exploit) diversity in the individual predictors. That is, if the individual predictors exhibit different patterns of generalization, then the strengths of each of the predictors can be combined to form a single stronger predictor. A lot of research in ensemble learning has gone into finding methods that encourage diversity in the predictors.

Dietterich [10] suggests three reasons why ensembles perform better than the individual predictors. The first reason is statistical. A learning algorithm can be considered as a way to search the space of hypotheses to identify the best hypothesis in it. The statistical problem is caused due to insufficient data. Due to this problem, the learning algorithm would give a set of different hypotheses with similar accuracy on the training data. By ensembling them, the risk of choosing the wrong hypothesis would be averaged out. The second reason is computational. Often, while looking for the best hypothesis, the algorithm might be stuck in local optima, thus giving us a bad hypothesis. By considering multiple such hypotheses, we can obtain a much better approximation to the true function. An example of the computational aspect is trying to train a neural network by restarting gradient descent a number of times to ensure that the result is better. The third reason is representational. Sometimes the true function might not be any hypothesis in the hypotheses space. By ensembling them, the representational space might be expanded to give a better approximation of the true function. Given the discussion about ensemble methods, we now consider combining the predictions in the method in section III-A.

*Methodology for Combining Predictions*

With each prediction model having access to different information about the data, combining them improves the representation and averages out the chance of finding an improper hypothesis. Hence we expect a combination to give an improvement in accuracy. As an example for improving representation, suppose a linear regression is to be used for training on the dataset. Such an arrangement will likely have a high bias on a real world dataset. Using linear regression on the clusters and not on the entire dataset gives a chance to expand the representational space and give a better fit to the data and increase variance.

As discussed in section III-A, we obtain a set of K predictions by varying the value the free parameter 'k'. These predictions can be combined by uniform averaging, weighted averaging or ensembling them together. The aim of our work is to show the utility of clustering in causing an improvement in accuracy, and hence though we can use ensemble methods to combine them together we show results by simple averaging only. Averaging the predictions in a regression task (equivalent to voting in a classification task) is probably the easiest way to combine them. First, the training set is clustered and by varying k, K prediction models are obtained. And then each of these prediction models are used to make a prediction on the test set. We thus obtain a set of K predictions on the test set. Averaging all these predictions might not be fruitful as some of them might be poor predictors and thus might prove to be detrimental to the prediction accuracy. Thus, a subset of the total number of predictions obtained must be averaged to improve accuracy. Like mentioned earlier, in place of uniform averaging, a weighted averaging or the use of an ensemble method could greatly improve the combined prediction.

### *C. Similarity with Other Existing Methods*

Before looking at the empirical evaluation of the method so discussed, we compare this method with some papers that atleast talked of using clustering for prediction.

We introduced a simple yet effective bootstrap-aggregating meta-algorithm that uses clustering as means to bootstrap. This method can be thought of as a mixture of local experts similar to one discussed by Jacobs, Hinton *et al.* [13]. It is noteworthy however that unlike in other bagging methods which select a random subset of the data to bootstrap, this method has a specific expert for each "locality" i.e cluster; which can potentially lead to more interpretability. By varying the granularity of the clustering we are able to train a set of experts at different scales which leads to a set of diverse predictions amenable to ensembling together. For example, if a K of 10 is chosen then for each test point there are ten experts to "consult" for a prediction, one each at a different level of granularity (i.e for k =1 there is an expert, at k =2 there is another and so on till k = 10).

On their work on Statistical Predicate Invention, Kok & Domingos [14] use multiple clusterings to better capture the interactions between objects in relational learning. Deodhar & Ghosh [15] also mention the same, however they use it in co-clustering framework and both of these works do not combine the predictions at different scales.

## IV. Empirical Validation

In this section we report the mechanics of an empirical study performed on a number of benchmark datasets for the task of regression for three different predictors.

### *A. Algorithms*

The algorithm used for clustering the various datasets was the k-means algorithm. k-means finds a partition by optimizing a distortion function, and while it can be considered to converge in a certain sense (it can be stated to be Lyapunov Stable [16] and thus the objective function decreases monotonically), the distortion function for k-means is non-convex. It is thus sensitive to the choice of initial cluster centroids and returns sub-optimal solutions quite often. We randomly initialized k-means 200 times on each run and picked the best solution.

For the prediction task (i.e. for training cluster models), Linear Regression, Step-Wise Linear Regression and Random Forests (for regression) were used. While this work can be extended to classification tasks as well, we do not discuss them in this work.

### *B. Datasets*

The datasets used for the empirical validation of this technique were taken from the University of California, Irvine Machine Learning repository [17]. Out of the 17 regression datasets available, those datasets were considered that did not have a large number of missing values or nominal attributes and thus we restricted ourselves to datasets having numerical attributes solely. Some of these datasets had more than one target variable. Such cases are reported as separate datasets.

The following datasets were considered (a) Breast Cancer Wisconsin Dataset (BREAST CANCER) has 569 data instances, each having 32 attributes. The prediction is for diagnosis (Benign or Malignant); (b) Cement Compressive Strength Dataset (COMPRESSIVE) has 1030 data points in total. Each data instance is described by 10 features [18]. The task is to predict the compressive strength (M Pa); (c) Concrete Slump I; (d) Concrete Slump II and (e) Concrete Slump III are essentially the same dataset (CONCRETE SLUMP) with the target attribute different in each case. This dataset has 103 data instances and 10 attributes, out of which 3 are target attributes (slump, flow and compressive strength); (f) The Forest Fires Dataset (FIRES) is one of the hardest regression datasets available[19]. It has 13 attributes and a total size of 513 observations. The task is to predict area burned in square kilometers; (g) Housing Dataset (HOUSING) has 506 instances of houses around the suburbs of Boston. There are 14 attributes; the task is to predict the median value of owner occupied houses in $1000's. The Parkinson's Telemonitoring Dataset (PARKINSON) [20] is a unique dataset in which about 5875 instances are provided, each with 26 attributes. This dataset has two target attributes which we denote as (h) Parkinson – I and (i) Parkinson – II; (j) Red Wine and (k) White Wine are two extensive datasets [21] that have 1599 and 4898 data points respectively, each with 12 features. Out of which one, the wine quality score (between 0 and 10) is the target attribute. These datasets give us a desired variety to test empirically our approach. Some of these datasets are straightforward tasks, while some are (such as FIRES) are amongst the hardest regression datasets available.

### *C. Methodology*

For testing the efficacy of this method, the datasets were subject to a 5 fold cross validation. No feature selection was done on any of the datasets. This is beneficial in these experiments as that makes the prediction task harder. Some of these datasets have a large number of attributes and hence it is clear that not doing feature selection would make the prediction task harder. The only dataset in which a set of features were chosen was the forest fires dataset (f). As given in the description of the dataset in the UCI Machine Learning Repository, we used the last four attributes only.

Features in all datasets were also normalized to values between 1 and -1 before applying this technique. This normalization was simply to ensure that none of the features dominated disproportionately in the clustering or regression tasks. While other normalization procedures were tested and some datasets returned better results with specific normalization techniques, we report the results only with one technique applied uniformly across datasets.

Two methodologies for combining predictions were employed in the experiments. Following is one of them:.

1. Normalize the dataset such that the features are scaled to the interval [-1, 1]
2. Run k-means clustering on the dataset from 2 to k and assign the value of k for which the dataset hit an empty cluster ($K_{empty}$) to it.
3. Choose K = $K_{empty}/2$ for that dataset. This will signify how many prediction models are to be obtained. Clearly, $K_{empty}/2$ prediction models (discussed in section III) are obtained.
4. For each prediction model obtained in step 3 obtain a prediction on the test set.
5. Uniformly average all predictions in step 4 to get a final prediction.

Clearly this method is simplistic in choosing a fixed value of k and not choosing a value empirically. To offset this problem we use a second methodology too. This is described below:

1. Normalize the features between [-1, 1] like in the previous case.
2. Recall that we have to run a 5 fold cross validation on the data. In each of the 5 runs, we have randomly chosen and mutually exclusive train and test sets, such that $4/5^{th}$ of the data forms the train set and the remaining $1/5^{th}$ forms the test set.
3. For each of the five folds, run a sub – 5 fold cross validation on the training data of that fold (a cross validation within a cross validation i.e consider the $4/5^{th}$ of the data mentioned in step 2 and divide it further into 5 folds).
    a. In each such sub cross validation phase consider a high value of k (such as $K_{empty}$) and

cluster the training data of this sub phase till that value.

   b. Train prediction models till this value of k in the sub-phase training set or to a value of k where models can be trained.
   c. Average the predictions obtained in this 5 fold sub-cross validation from prediction models 1 to the value in step 3 b above.
   d. Find the k in step c averaging to which (from $PM_1$ to $PM_k$) gives the least prediction error
   e. Choose this k and return it to the main cross validation loop
4. The k returned in step 3 e. is the value for that fold to which the predictions are to be averaged to. i.e. train prediction models on the train set to this value of k and average the predictions of all of these prediction models.
5. Repeat the process for each fold.
6. Average the errors in the five folds to get a single prediction error (let's call it CVk error)

As discussed, the problem with the first method was that no matter what predictor was used, it always averaged the first $K_{empty}/2$ prediction models. This value did not depend on what predictor was used to make the final prediction. Clearly the choice of predictor would have an impact on how many prediction models are to be averaged (intuitively a weaker predictor would need more prediction models to improve performance while a stronger one would need fewer). The second method alleviates this problem to some degree. It however suffers from the problem that training in the sub-cross validation phase, by virtue of having lesser points than training in the cross validation phase might return prediction models that are not completely representative of the prediction models returned in the main cross validation.

With these methodologies, experiments were run using three predictors:

1. Linear Regression (without feature selection)
2. Stepwise Linear Regression
3. Random Forests (for regression)

There were multiple objectives to the experiments conducted using these two methodologies, some of which were:

1. In what kind of datasets is such a method of averaging predictions useful? Are there datasets when it does worse?
2. The choice of averaging $K_{empty}/2$ predictors is an approximation. However it would be interesting to see how the value of number of prediction models that returns the best error value changes depending on the nature of the dataset and the predictor.
3. How much does the utility of clustering depend on the predictor used? What if a strong predictor is used and what if a weak predictor is used?
4. How do these results compare with results when a cross validation within a cross validation is used to choose a value of k till which to average.
5. Does the nature of data normalization alter results?

## V. Results

The three different predictors (Linear Regression, Stepwise Linear Regression and Random Forests) were chosen as representatives for different levels of predictor complexity. A linear regression model might be considered to have high bias with respect to most real world datasets and hence might be thought of as a naïve choice in most prediction settings. Stepwise Linear Regression on the other hand usually does a better job than its forced counterpart. Random Forests, however, represent the state of art in classification and regression. As discussed in the previous section, the experiments were done so as to evaluate how the information exploited by clustering the data aided in a prediction task given a dataset and type of predictor used. Another important question was to understand what kinds of datasets were suitable for such a technique. These observations are discussed in this section. The results with clustering are compared to the condition when no clustering was used ($PM_1$) using a paired t-test to check for statistical significance. The two methodologies described in section IV-C were employed to combine predictions.

The results are organized in three tables (Tables I, II and III) and two figures (Figs 4 and 5). The prediction results with clustering (employing both the methodologies discussed in IV C) and without clustering ($PM_1$) for Linear Regression, Stepwise Linear Regression and Random Forests are tabulated in tables I, II and III respectively. Figures 4 and 5 show the error profiles for the different datasets for Stepwise Linear Regression and Random Forests. The error profile shown is essentially the mean absolute error in the prediction obtained by ensembles having 2 to k prediction models (this is represented in the x axis. i.e. a point 5 on the x axis would mean that the bar graph at that point shows the error returned by a model that averaged the first five prediction models). These figures underline the fact that the choice of k returned using the first methodology (of taking $K_{empty}/2$) gives quite a sub-optimal choice of k and thus the error value. While the CVk error (given by the second methodology for choosing k empirically) cannot be plotted in such graphs for obvious reasons, the number in the tables show a marked improvement over the first methodology.

In table I, we immediately notice a couple of broad trends: The CVk error is mostly better than the error obtained by averaging the first $K_{empty}/2$ prediction models (as indicated by kmeans – I i.e. methodology I in the table). In all but one case it also improves the statistical significance for the improvement over $PM_1$. The only exception being the red wine dataset where the error returned by the second methodology CVk is a little worse than even $PM_1$, however this difference is not statistically significant. Perhaps this improvement across board is not surprising. This is because of the nature of the Linear Regression model, which is a very simple model that has a high bias w.r.t most real world datasets ($PM_1$). So clustering even a little and not to a level that is optimal (clearly methodology I chooses a k that clearly could have been better) improves the prediction accuracy significantly as it boosts the variance. Improving this estimate of how many prediction models should be averaged (by using methodology II, CVk) further improves the prediction accuracy and statistical significance over $PM_1$. Another observation was that there are datasets which are more clusterable than others with respect to the size of the data matrix (rows by columns or number of data points by number of features). In such datasets, the improvement in prediction errors is not only huge, it is highly statistically significant. The only exception to this generalization is the Red Wine dataset. The red wine dataset is a moderate sized dataset as compared to the others, however clustering does not seem to help in prediction with it. In another dataset, Slump II, the prediction made by using methodology I is better than $PM_1$ but only marginally statistically significant, this improvement is made statistically significant by using the second methodology. This underlines the ability of CVk to find a better prediction. In conclusion, out of the 11 datasets, an improvement in prediction accuracy was seen in all of them (except the CVk error for Red Wine), this improvement was much more pronounced in the CVk error, both in terms of raw error and statistical significance (over $PM_1$). This observation points out that the choice of k to average in method I was perhaps suboptimal. This method of choosing k itself might not be optimal but certainly is more principled than the method employed in some experiments.

TABLE I

PREDICTIONS USING LINEAR REGRESSION AND CLUSTERING

| Dataset | $K_{empty}$ | MAE ($PM_1$) | MAE (kmeans - I) | p-value(with $PM_1$) | MAE- CVk | p-value(with $PM_1$) |
|---|---|---|---|---|---|---|
| Parkinson I | 42 | 6.3445 | 5.0809 | << 0.001 | 4.3638 | << 0.001 |
| Parkinson II | 42 | 8.0785 | 6.6190 | << 0.001 | 5.7727 | << 0.001 |
| Red Wine | 26 | 0.5065 | 0.5048 | 0.6860 | 0.5073 | 0.7888 |
| White Wine | 52 | 0.5858 | 0.5507 | << 0.001 | 0.5394 | << 0.001 |
| Housing | 35 | 3.4021 | 2.5904 | << 0.001 | 2.5883 | << 0.001 |
| Breast Cancer | 20 | 0.1944 | 0.1136 | << 0.001 | 0.1139 | << 0.001 |
| Fires | 11 | 19.5009 | 18.9246 | 0.0399 | 18.8739 | 0.0074 |
| Concrete | 30 | 8.2730 | 5.9688 | << 0.001 | 5.8316 | << 0.001 |
| Slump I | 7 | 6.2958 | 5.8312 | 0.0959 | 5.7297 | 0.0155 |
| Slump II | 7 | 11.2000 | 10.5712 | 0.1843 | 10.5203 | 0.1343 |
| Slump III | 7 | 2.0136 | 1.7655 | 0.0086 | 1.7475 | 0.0063 |

$K_{emoty}$ is the value of k in k-means beyond which the dataset returned empty clusters. kmeans – I represents the first heuristic where $K_{empty}/2$ prediction models were averaged to get a prediction. The errors are reported in Mean Absolute Error, along with the ttest values with $PM_1$ reported. CVk reports the prediction errors when the second heuristic is used, along with the p-values when compared to $PM_1$.

TABLE II

PREDICTIONS USING STEPWISE LINEAR REGRESSION AND CLUSTERING

| Dataset | $K_{empty}$ | MAE ($PM_1$) | MAE (kmeans - I) | p-value(with $PM_1$) | MAE- CVk | p-value(with $PM_1$) |
|---|---|---|---|---|---|---|
| Parkinson I | 42 | 6.3597 | 5.1411 | << 0.001 | 4.4290 | << 0.001 |
| Parkinson II | 42 | 8.0798 | 6.7266 | << 0.001 | 5.8678 | << 0.001 |
| Red Wine | 26 | 0.5059 | 0.5034 | 0.4799 | 0.5005 | 0.1555 |
| White Wine | 52 | 0.5850 | 0.5537 | << 0.001 | 0.5440 | << 0.001 |
| Housing | 35 | 3.4252 | 2.5403 | << 0.001 | 2.5503 | << 0.001 |
| Breast Cancer | 20 | 0.1962 | 0.0941 | << 0.001 | 0.0784 | << 0.001 |
| Fires | 11 | 19.2495 | 18.8972 | 0.0215 | 18.8314 | 0.0368 |
| Concrete | 30 | 8.3243 | 6.1101 | << 0.001 | 5.8025 | << 0.001 |
| Slump I | 7 | 6.4607 | 6.1754 | 0.2709 | 5.7699 | 0.0255 |
| Slump II | 7 | 10.5918 | 10.6652 | 0.8909 | 10.5639 | 0.8376 |
| Slump III | 7 | 2.1864 | 1.7687 | << 0.001 | 1.7880 | << 0.001 |

TABLE III

PREDICTIONS USING RANDOM FORESTS AND CLUSTERING

| Dataset | $K_{empty}$ | MAE ($PM_1$) | MAE (kmeans - I) | p-value(with $PM_1$) | MAE- CVk | p-value(with $PM_1$) |
|---|---|---|---|---|---|---|
| Parkinson I | 42 | 2.0790 | 2.0687 | 0.5650 | 1.8468 | << 0.001 |
| Parkinson II | 42 | 2.6942 | 2.6900 | 0.8363 | 2.4264 | << 0.001 |
| Red Wine | 26 | 0.4233 | 0.4255 | 0.4239 | 0.4211 | 0.3260 |
| White Wine | 52 | 0.4312 | 0.4290 | 0.2313 | 0.4297 | 0.3186 |
| Housing | 35 | 2.1888 | 2.2046 | 0.7394 | 2.1764 | 0.6789 |
| Breast Cancer | 20 | 0.0777 | 0.0760 | 0.4359 | 0.0760 | 0.3540 |
| Fires | 11 | 20.6546 | 20.5958 | 0.8879 | 20.1743 | 0.0490 |
| Concrete | 30 | 3.5550 | 3.7846 | << 0.001 | 3.5202 | 0.0489 |
| Slump I | 7 | 5.5629 | 5.8020 | 0.1086 | 5.7336 | 0.3031 |
| Slump II | 7 | 10.1521 | 10.1762 | 0.9537 | 10.1521 | -- |
| Slump III | 7 | 3.1424 | 2.9877 | 0.0390 | 3.0051 | 0.0373 |

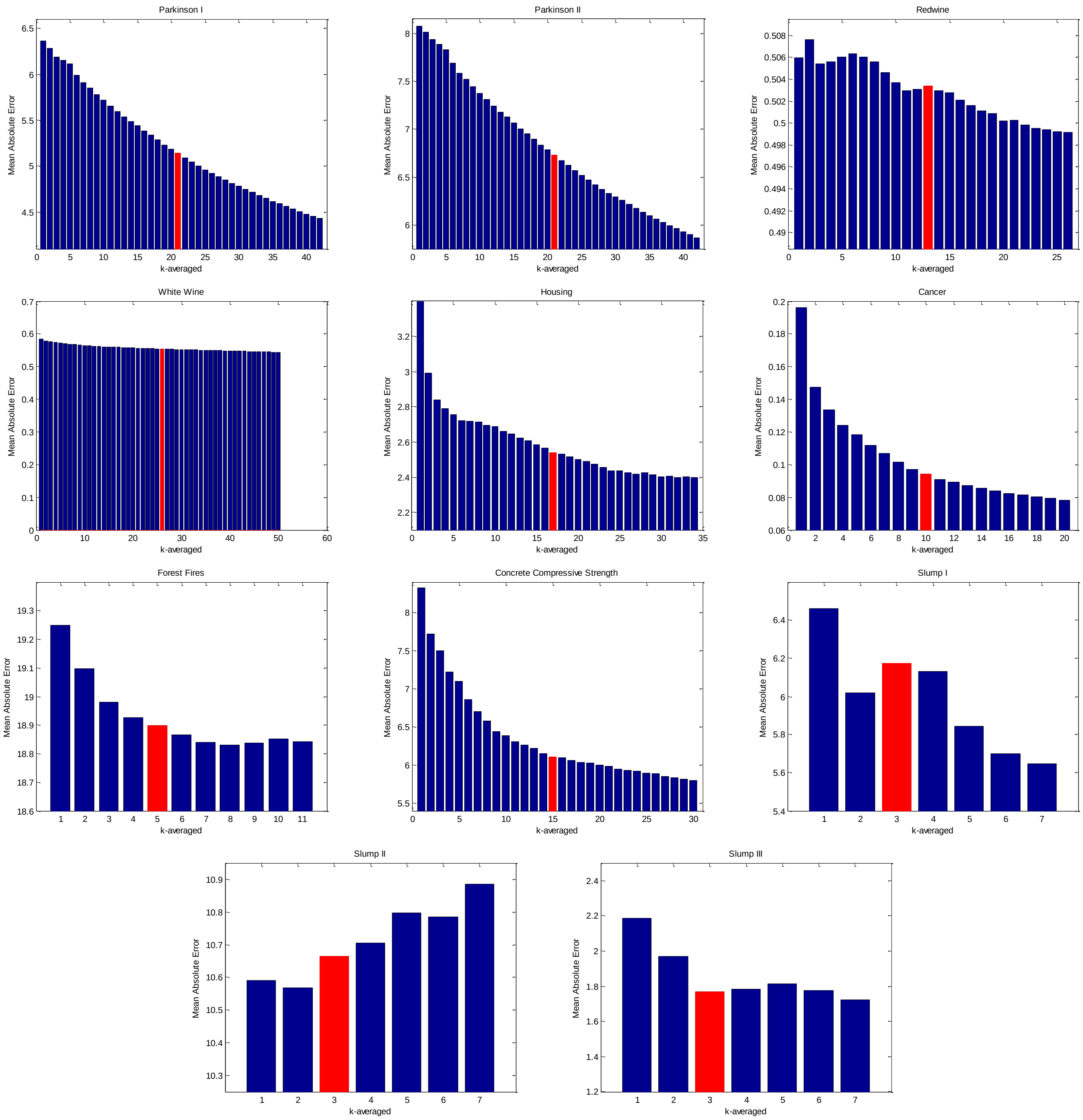

Fig. 4. The error profiles for all 11 datasets for stepwise linear regression. The x-axis represents the number of prediction models averaged from 1. The bar marked in red indicates the one that has been chosen by the first heuristic as the final prediction. In many cases we notice that this is clearly a sub-optimal choice. The chosen value and the lowest value in the error profile for each dataset should be contrasted with the value of CVk mentioned in the table. Since the number of prediction models to average chosen is different in each fold by the second method, it has not been represented in the graph.

Table 2, which aggregates the results for Stepwise Linear Regression, shows trends similar to Linear Regression. The CVk errors are generally better as compared to the errors returned by the first methodology here as well. The only two exceptions in which clustering (by both methods) does not seem to improve upon $PM_1$ are the SLUMP II and Red Wine datasets (just like for linear regression). As expected, results for stepwise linear regression with clustering give smaller errors as compared to simply linear regression with clustering. Like in the case of Linear Regression, the choice of the number of prediction models to average was suboptimal. This is indicated by the error profile (Fig. 4) for all 11 datasets when stepwise linear regression was used. The bar in the graph marked in red indicates the error and k picked by using the

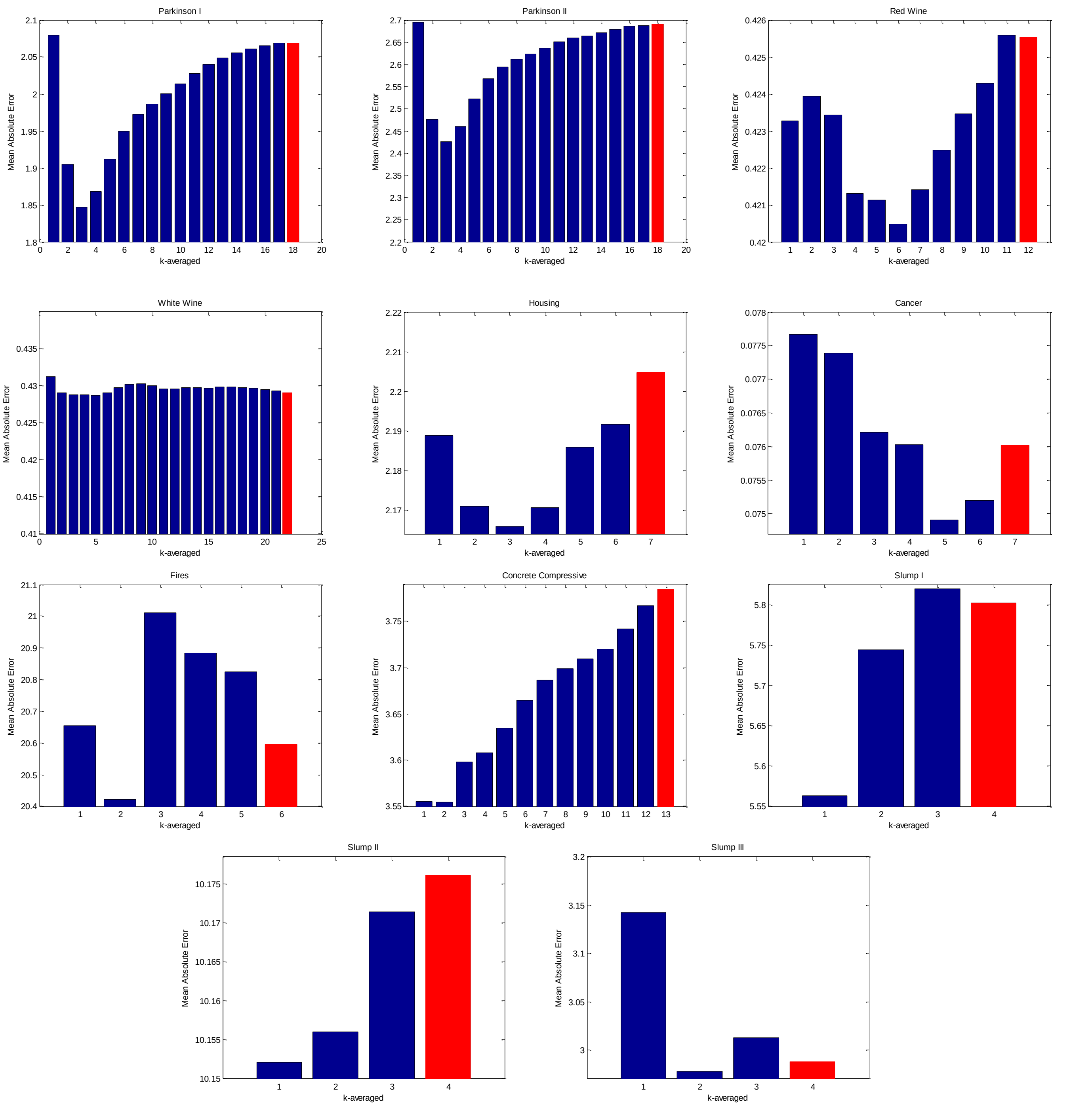

Fig. 5. The error profiles for all 11 datasets for Random Forests (for regression). The x-axis represents the number of prediction models averaged from 1. The bar marked in red indicates the one that has been chosen by the first heuristic as the final prediction. In many cases we notice that this is clearly a sub-optimal choice. The chosen value and the lowest value in the error profile for each dataset should be contrasted with the value of CVk mentioned in the table. Since the number of prediction models to average chosen is different in each fold by the second method, it has not been represented in the graph.

first heuristic. These can be contrasted with the CVk errors. The error profiles make a strong case for choosing the number of prediction models to average empirically. Similar error profiles were observed for Linear Regression. Since the two methods are simple in terms of representation power, more clustering seems to help the results (making $K_{empty}/2$ a bad choice), this is especially prominent in datasets that are more clusterable (such Parkinson I and II, White Wine etc), red wine being the only exception. This notion is also reinforced by the error profiles in smaller, noisier datasets such as the SLUMP datasets. By choosing k empirically, we frequently choose a better k for each fold and this is reflected in the results.

The results for random forests are the most interesting. This is because it is a strong predictor by itself and hence it is not clear how much help clustering would lend to improve prediction accuracy. It being a strong predictor also in turn means that the earlier heuristic (first methodology) of choosing how many prediction models to average would not work.

Choosing this number empirically seems to be a better bet (CVk). This is reflected in the error profiles for all the datasets (Fig. 5), which are very different from the error profiles of simpler predictors such as Linear Regression and Stepwise Linear Regression. We also notice that the red bar is usually much worse in terms of results. Also, the "correct" choice of how many prediction models to average seems to change from dataset to dataset and there does not seem to be a clear trend unlike for linear regression and stepwise linear regression. Table III has results that confirm the above speculations. Except in a couple of datasets, the first methodology for combining predictions does not help in improving the prediction accuracy at all. In fact, it goes worse in more than half of the datasets and significantly worse in one dataset. The results for CVk as expected are much better; with the prediction errors improving across datasets and importantly, significantly improving over $PM_1$ (Random Forest on the entire dataset with no clustering) in 6 datasets. This is an important result. Even in the dataset where the first method returned a significantly worse prediction, the CVk error is better, though not statistically significant. As a remark on implementation, it should be noted that Random Forests could not be trained to a high enough value of k as they need a certain number of points to train properly. And hence much lesser values of k are shown in the bar graphs beyond which training Random Forests was untenable.

One of the advantages of choosing k empirically is illustrated very clearly in the case of SLUMP II. In this dataset, clustering does not seem to give any advantage in prediction at all. The cross validation within cross validation affirms this and returns the best value of k to be 1. This means that we end up with a final prediction which is the same as for $PM_1$. This example shows that choosing k empirically ensures that we do not force clustering on a dataset where its performance after clustering will actually go worse.

## VI. Discussion and Future Work

The results obtained in using clustering in conjunction with Linear Regression are not very surprising. The Linear Regression Model is a model with a high bias and is thus not expected to do too well on most real world datasets. Using Linear Regression in conjunction with clustering makes it a much more powerful method as it gives it access to more variance in the data, thus improving the bias-variance trade-off of the complete system. The improvement in prediction accuracy is very significant when it is combined with clustering after doing some feature selection (stepwise regression). In some cases stepwise with clustering returned accuracies comparable to those returned by Random Forests without clustering. Therefore, clustering seems to be giving a cheap method of accessing a lot of information about the data.

It must also be noted that clustering a dataset at a single value of k, with any predictor (only one PM alone making a prediction without any ensembling), rarely improved prediction accuracy in a statistically significant manner compared to the predictor trained without clustering. But, if done at different scales with a prediction obtained at each scale and then combined by means of a naïve ensemble, the improvement is very significant as discussed in the previous section. It was also observed that in datasets that were not very clusterable, this technique did not improve upon much.

The experiments done using random forests were more interesting. On smaller datasets the results obtained by using a random forest on the entire dataset and those obtained using the combination of predictions obtained at different scales of clustering did not have a statistically significant difference. This is understandable, as for small datasets clustering at a high value of k might not be able to reveal the true structure for lack of enough data points and might just end up considering sampling noise as structure [3]. This would not contribute much information to aid in the prediction task (might instead reduce the quality). The second and the more important reason would be that for small datasets, techniques such as random forests can exploit enough information such that the generalization error on the test set approaches a limit. Since Random Forest is itself an ensemble method, by means of random sampling of instances and attributes, it already gains a lot of information about the data. Because of this reason, information provided by clustering might not be necessarily novel. An implicit justification for this is given by the results returned by datasets that are large in size and are much more clusterable. Clustering in such cases is thus more likely to give a novel source of variance that can improve prediction significantly.

An important aspect about the method was choosing which predictions to average. One of the methodologies followed was a naïve averaging of the first half of the predictions. This was a suboptimal choice, as there could have been better combinations of the set of predictions that could have been averaged. The choice of using the first half of the predictors was based on the following intuition: Finding the optimal clustering for a dataset might also be considered to be a bias-variance problem. If the number of clusters is too few as compared to the "true" number of clusters, then, most likely, the clustering has a high bias. Inversely, if the number of clusters is too high, we would be over fitting on the data. We selected the first half as a crude tradeoff between this tension. Ideally, the optimal choice of the predictors would be a function of both the clusterabilty of the dataset and the base predictor used. For example, if Linear Regression is used, averaging more predictions could be beneficial. The point of the method discussed in this work was to indicate that clustering gives access to a novel source of information in the data, and thus the aspect of combinations was not optimized. However, a method to pick k empirically was still employed and experimented with. It showed superior results to the earlier naïve heuristic. There were some problems with this methodology too. One being that k-means clustering is not a particularly stable clustering. The method utilized to choose a k was based on a cross validation within each fold. Since this stage chose a sub fold that was of a smaller size that the original fold it was not necessarily representative of it. And thus many times it was observed that the error profiles for the sub-cross validation phase were quite different from the error profiles for the main cross validation phase (Fig. 4 and Fig. 5 have the error profiles of the main cross validation phase). While this definitely hurt the best choice of k, this experiment establishes how the prediction could be improved. This discussion poses an open model selection problem that could be solved by methods such as those used by the authors in the KDD cup [22] or using averaging as discussed by Caruana [23].

Perhaps the best method for model selection in this case would be the PAC-MDL bound [8].

Another open question is if injecting randomness at various stages can improve the method's prediction performance. This randomness can be injected in many stages, such as: Currently, we assign each test point to a cluster centroid based on the Euclidean distance and make a prediction for that point. Instead, the point could be assigned in a fuzzy manner, with probabilities of it lying in all clusters. Predictions on each cluster can then be obtained for that point and then weighted averaging can be done to obtain the final prediction. The weights in this case would be the probability that the point belongs to a particular cluster.

Also, for each cluster model we use all features and training examples in the cluster. A random selection with replacement can be made to generate more diversity in the predictors. Preliminary work shows that such an ensemble gives promising performance. Yet another source of variance can be the k-means clustering algorithm itself. The k-means algorithm can give unstable results. In the experiments, we ran kmeans 200 times and picked the best clustering. However, each of the converged runs can be used to generate more predictions that can then be combined together.

Yet another area that can be worked on to improve the performance of the system can be by using supervised clustering. In our task, we use clustering to boost a prediction performance. However, the clustering is done in a completely unsupervised manner without any regard to the target. The clustering might be completely different if the target is accounted for. A process where the target is taken into consideration while clustering and then models are trained on these clusters would potentially be more beneficial.

## Acknowledgment

We thank Dr. Carolina Ruiz, Dr Sergio Alvarez and Dr. Alexandru Niculescu-Mizil for helpful suggestions and discussions about the work.